\documentclass[11pt,a4paper]{article}
\usepackage{acl2023} 
\usepackage{times}
\usepackage{latexsym}

\usepackage{amsmath}
\usepackage{amsfonts}
\usepackage[linesnumbered,ruled,vlined]{algorithm2e}
\usepackage{graphicx} % DO NOT CHANGE THIS
\usepackage{booktabs}
\usepackage{array}
\usepackage{enumitem}
\usepackage{amsthm}
\usepackage{algpseudocode}
\usepackage[switch]{lineno}
\usepackage[utf8]{inputenc}
\usepackage{eqparbox}
\usepackage[nopar]{lipsum}
\usepackage{multirow}
\usepackage{makecell}
\usepackage{xcolor}
\usepackage{hhline} 
\usepackage{microtype}
\usepackage{diagbox}

\usepackage{adjustbox}

\usepackage{relsize}

\usepackage{booktabs,multirow,array}
\newcolumntype{N}{@{}m{0pt}@{}}%a fix for array package

\usepackage[many]{tcolorbox}
\newtcolorbox{fancyquotes}{%
    enhanced jigsaw, 
    breakable,      % allow page breaks
    frame hidden,   % hide the default frame
    left=0.5cm,       % left margin
    right=0.1cm,      % right margin
    overlay={%
        \node [scale=8,
            text=black,
            inner sep=0pt,] at ([xshift=-1cm,yshift=-1cm]frame.north west){}; 
        \node [scale=8,
            text=black,
            inner sep=0pt,] at ([xshift=1cm]frame.south east){};  
            },
                parbox=false,
}

\usepackage{mathtools, nccmath}
\usepackage{scrextend}
\deffootnote[.25in]{.25in}{.15in}{\makebox[.25in][r]{\thefootnotemark .\hspace{.15in}}}

\makeatletter

\newtheorem*{proof*}{Proof}

\newcommand{\red}[1]{\textcolor{red}{#1}}
\newcommand{\blue}[1]{\textcolor{blue}{#1}}

\usepackage{listings}
\usepackage{color}
\definecolor{codegreen}{rgb}{0.3,0.5,0.0}
\def\@fnsymbol#1{\ensuremath{\ifcase#1\or \dagger\or *\or \ddagger\or
   \mathsection\or \mathparagraph\or \|\or **\or \dagger\dagger
   \or \ddagger\ddagger \else\@ctrerr\fi}}

\newcolumntype{C}[1]{>{\centering\let\newline\\\arraybackslash\hspace{0pt}}m{#1}}

\ExplSyntaxOn
\NewExpandableDocumentCommand { \ValuePlusOne } { m } 
  { \int_eval:n { \int_use:c { c @ #1 } + 1 } }
\NewExpandableDocumentCommand { \Sec } { m } 
  { \fp_eval:n { secd ( #1 ) } }
\NewDocumentCommand { \Rot } { m }
  { 
    \hbox_to_wd:nn { 1 em }
      { 
        \hbox_overlap_right:n 
          { 
            \skip_horizontal:n { \fp_to_dim:n { 7 * cosd (\Angle) } } 
            \rotatebox{\Angle}{#1}
          } 
      } 
  }
\ExplSyntaxOff

\def\Angle{45}
    
\bigskip
\def\Angle{90}

\title{RetroMAE-2: Duplex Masked Auto-Encoder For Pre-Training Retrieval-Oriented Language Models} 

\author{Shitao Xiao$^1$\thanks{The two people contribute equally to this work and are designated as co-first authors.}, Zheng Liu$^2$\footnotemark[1], Yingxia Shao$^1$, Zhao Cao$^2$ \\
  1: Beijing University of Posts and Telecommunications, Beijing, China \\ 
  2: Huawei Technologies Ltd. Co., Shenzhen, China \\
  \texttt{stxiao@bupt.edu.cn}, 
  \texttt{zhengliu1026@gmail.com}
}

\begin{document}
\maketitle 

%% 
%% pre-trained language models
%% retrieval oriented pre-training
%% masked-auto encoder and retromae
%% only rely on [cls]
%% ordinary tokens are also important
%% -- our work ---
%% propose duplex masked auto-encoder
%% jointly pre-train representation capability of [cls] and ordinary tokens

\begin{abstract}
% Pre-trained language models are widely used as the backbone networks for deep semantic retrieval. To better support this application, growing attention is paid to developing retrieval-oriented language models \cite{liu2022retromae,wang2021tsdae,gao2021condenser}. 
To better support information retrieval tasks such as web search and open-domain question answering, growing effort is made to develop retrieval-oriented language models, e.g., RetroMAE \cite{liu2022retromae} and many others \cite{gao2021condenser,wang2021tsdae}. 
Most of the existing works focus on improving the semantic representation capability for the contextualized embedding of the [CLS] token. However, recent study shows that the ordinary tokens besides [CLS] may provide extra information, which help to produce a better representation effect \cite{lin2022aggretriever}. As such, it's necessary to extend the current methods where all contextualized embeddings can be jointly pre-trained for the retrieval tasks.

In this work, we propose a novel pre-training method called Duplex Masked Auto-Encoder, \textit{a.k.a.} DupMAE. It is designed to improve the quality of semantic representation where all contextualized embeddings of the pre-trained model can be leveraged. It takes advantage of two complementary auto-encoding tasks: one reconstructs the input sentence on top of the [CLS] embedding; the other one predicts the bag-of-words feature of the input sentence based on the ordinary tokens' embeddings. The two tasks are jointly conducted to train a unified encoder, where the whole contextualized embeddings are aggregated in a compact way to produce the final semantic representation. DupMAE is simple but empirically competitive: it substantially improves the pre-trained model's representation capability and transferability, where superior retrieval performances can be achieved on popular benchmarks, like MS MARCO and BEIR.

\end{abstract} 

\section{Introduction}

Neural retrieval is important to many real-world scenarios, such as web search, question answering, and conversational system \cite{huang2013learning,karpukhin2020dense,komeili2021internet,izacard2022few}. In recent years, pre-trained language models, e.g., BERT \cite{Devlin2019BERT}, RoBERTa \cite{Liu2019Roberta}, T5 \cite{raffel2019exploring}, are widely adopted as the retrievers' backbone networks. The generic pre-trained language models are not directly applicable to retrieval tasks. Thus, it calls for complex fine-tuning strategies, such as sophisticated negative sampling \cite{xiong2020approximate,qu2020rocketqa}, knowledge distillation \cite{hofstatter2021efficiently,lu2022ernie}, and the joint optimization of retriever and ranker \cite{ren2021rocketqav2,zhang2021adversarial}. To reduce this effort and bring in better retrieval quality, there are growing interests in developing retrieval-oriented language models. One common practice is to leverage self-contrastive learning \cite{chang2020pre,guu2020realm}, where the language models are learned to discriminate heuristically acquired positive and negative samples in the embedding space. Later on, auto-encoding is found to be more effective \cite{wang2021tsdae,lu2021less}, where the language models are learned to reconstruct the input based on the generated embeddings. Recent works \cite{liu2022retromae,wang2022simlm} further extend the auto-encoding methods by introducing sophisticated encoding and decoding mechanisms, which brings about remarkable improvements of retrieval quality on a wide variety of benchmarks. 

% The existing retrieval-oriented pre-trained models mainly rely on the contextualized embedding from the special token, i.e., [CLS], to represent the semantic about input \cite{gao2021condenser,lu2021less,liu2022retromae,wang2022simlm}. However, recent study finds that other ordinary tokens may provide extra information and help to generate better semantic representations \cite{lin2022aggretriever}. Such a statement is consistent with previous research \cite{luan2021sparse,santhanam2021colbertv2}, as multi-vector or token-granularity representations may give higher discriminative power than those based on a single vector. Thus, it's necessary to extend the current works, where the representation capacity can be jointly enhanced for both [CLS] and ordinary tokens of the pre-trained model.  

The existing retrieval-oriented pre-trained models mainly rely on the contextualized embedding from the special token, i.e., [CLS], to represent the semantic about input \cite{gao2021condenser,lu2021less,liu2022retromae,wang2022simlm}. However, recent study finds that other ordinary tokens may provide extra information and help to generate better semantic representations \cite{lin2022aggretriever}. Such a statement is consistent with previous research \cite{luan2021sparse,santhanam2021colbertv2}, as multi-vector or token-granularity representations may give higher discriminative power than those based on one single vector. As a result, it is necessary to extend the previous works, such that the representation capability can be jointly pre-trained for both [CLS] and ordinary tokens. 

\begin{figure}[t]
\centering
\includegraphics[width=1.0\linewidth]{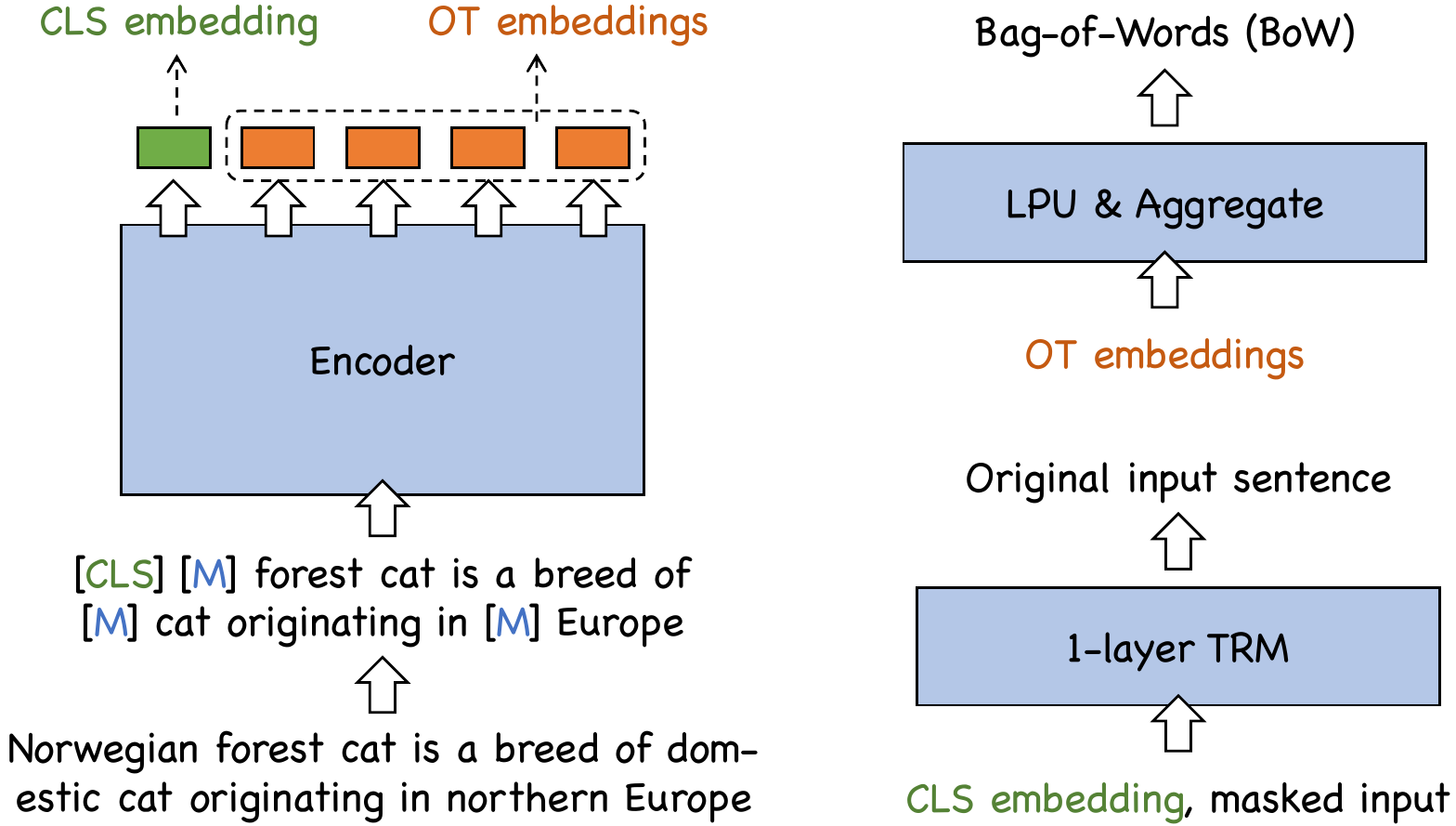}
% \vspace{-15pt}
\caption{\small{DupMAE. Encoder: the sentence is masked and encoded as the contextualized embeddings for [CLS] and ordinary tokens. Decoder: the CLS embedding is joined with the masked input, where the original input is recovered by an 1-layer transformer; OT embeddings are mapped into vocabulary space via LPU and aggregated to predict the BoW feature.}} 
\vspace{-10pt}
\label{fig:1}
\end{figure}

% With this motivation, we propose a novel pre-training method called the Duplex Masked Auto-Encoder, \textit{a.k.a.} \textbf{DupMAE}, shown as Figure \ref{fig:1}. 

To this end, we propose a novel auto-encoding framework called Duplex Masked Auto-Encoder, \textit{a.k.a.} \textbf{DupMAE} (Figure \ref{fig:1}). It employs two differentiated decoders working collaboratively, which aim to 1) improve each embedding's individual capacity, as well as 2) contribute to the quality of the joint representation derived from all embeddings.

$\bullet$ \textbf{Workflow}. DupMAE contains an unified encoder, which produces the contextualized embeddings for both [CLS] and ordinary tokens. The generated embeddings are used for two decoding tasks. On one hand, the [CLS] embedding, joined with the masked input, is used to recover the input sentence from an one-layer transformer. On the other hand, the ordinary tokens' embeddings are transformed into the vocabulary space (V), i.e, $|V|$-dim vectors, with a linear projection unit (LPU). The transformation results are aggregated into a $|V|$-dim vector by max-pooling, where the bag-of-words feature about the input is predicted.  

% $\bullet$ \textbf{Merits}. The above workflow is highlighted by its simple decoding process: an one layer transformer to recover the original sentence, and a linear projection unit to preserve the BoW feature. It brings two merits as a consequence. Firstly, the pre-training is made \texttt{Cost-Effective} given that all decoding takes operate at a low cost. Secondly and more importantly, the pre-training task is made highly \texttt{Demanding}: as the decoders are extremely simplified, it will force the encoding model to fully extract the input information such that high-fidelity reconstruction of the input can be produced.  

$\bullet$ \textbf{Merits}. The above workflow is highlighted by its simplicity: an one-layer transformer to recover the input, and a linear projection unit to preserve the BoW feature. Therefore, the pre-training is \texttt{Cost-Effective} given all decoding takes operate at a low cost. More importantly, the pre-training task is made highly \texttt{Demanding} on embedding quality: since the decoders are extremely simplified, it forces the encoder to fully extract the input information so that high-fidelity reconstruction can be made. Finally, the differentiated tasks may help the embeddings learn \texttt{Complementary} information: the [CLS] embedding focuses more on semantic information; while the OT embeddings, which directly preserve the BoW features, may incorporate more lexical information. 

% $\bullet$ \textbf{Representation}. The contextualized embeddings from [CLS] and ordinary tokens are aggregated in a simple way for the final representation. The [CLS] embedding is reduced to a lower dimension by linear projection. The ordinary tokens' embeddings, after transformed into the vocabulary space and aggregated by max-pooling, are sparsified by picking the top-N elements. The two results are concatenated as one vector. With properly configured linear projection and sparsification, it will preserve a similar memory footprint and cost of inner-product as the original dense embedding.  

$\bullet$ \textbf{Representation}. The contextualized embeddings from [CLS] and ordinary tokens are aggregated in a straightforward way to generate the representation of the input. The [CLS] embedding is reduced to a lower dimension by linear projection. The ordinary tokens' embeddings, after transformed into the vocabulary space and aggregated by max-pooling, are sparsified by selecting the top-N elements. The two results are concatenated as one vector. With a proper configuration of linear projection and sparsification, it may preserve the same memory footprint and cost of inner-product computation as the conventional methods.  

% Our proposed method is simple but empirically competitive. We make use of common pre-training data for DupMAE (Wikipedia, BookCorpus, MS MARCO), where a BERT-base scale encoder is learned. 
% Our empirical study verifies that the retrieval quality can be substantially improved for the pre-trained models thanks to the enhanced representation capability of the [CLS] and ordinary tokens' embeddings. Particularly, it achieves the state-of-the-art performances of a bi-encoder for both \textbf{MS MARCO} passage and document retrieval tasks, and gives rise to remarkable out-domain retrieval quality on \textbf{BEIR} benchmark. Our models and source code will be made publicly available. 

% Our proposed method is simple but empirically competitive. We perform DupMAE with common datasets: Wikipedia, BookCorpus, MS MARCO, where a BERT-base scale encoder is produced. According to our experiments, the pre-trained model helps to achieve remarkable performances across different scenarios: for \textbf{MS MARCO}, it reaches a MRR@10 of \textbf{42.6} on passage retrieval task and a MRR@100 of \textbf{45.1} on document retrieval task; for \textbf{BEIR}, it leads to an average NDCG@10 of \textbf{47.5} on all 18 datasets. Such results validate that the semantic representation capacities and transferability can be substantially improved thanks to DupMAE. Our models and source code will be made publicly available to facilitate the future research.  

Our proposed method is simple but empirically competitive. 
We perform DupMAE on common pre-training corpus where a BERT-based scale encoder is produced. Our pre-trained model achieves superior performances in various downstream tasks. For supervised evaluations on \textbf{MS MARCO}, it reaches a \underline{MRR@10 of {42.6}} in passage retrieval and a \underline{MRR@100 of {45.1}} in document retrieval. For zero-shot evaluations on \textbf{BEIR}, it achieves an average \underline{NDCG@10 of {49.1}} on all 18 datasets. It even notably outperforms strong baselines relying on more sophisticated fine-tuning approaches or much bigger model sizes. Thus, it validates that the representation capability and transferability of the pre-trained model can be substantially improved thanks to DupMAE. 
To facilitate future research and real-world applications, our model and source code will be made publicly available. 

\begin{figure*}[t]
\centering
\includegraphics[width=1.0\textwidth]{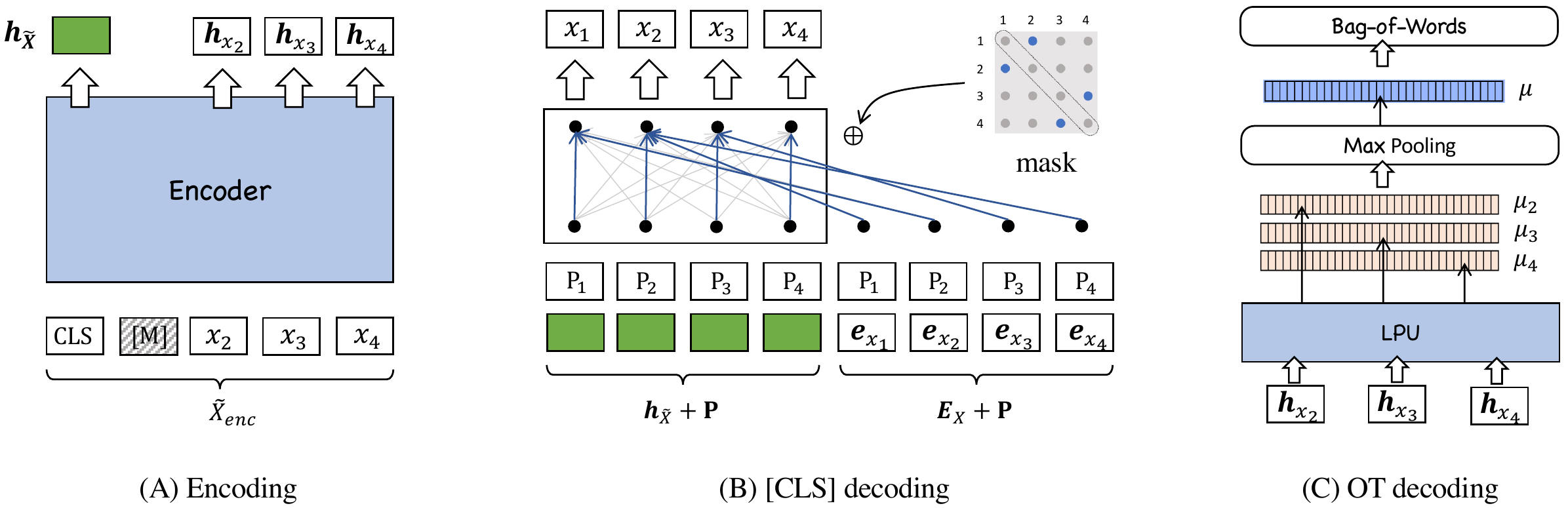}
\vspace{-18pt}
\caption{\small{Framework of DupMAE. The unified encoder generates the contextualized embeddings for the [CLS] and ordinary tokens (OT). The [CLS] decoding reconstructs the original sentence leveraging  an one-layer transformer; the OT decoding predicts the BoW feature of the input on top of the linear projection unit (LPU) and max-pooling.}}   
\vspace{-6pt}
\label{fig:2}
\end{figure*} 

% The framework of DupMAE is shown as Figure \ref{fig:2}. There is an unified encoder (A), where the masked sentence is encoded into the contextualized embeddings. There are two decoding components. One is for [CLS] decoding (B): it is based on a single-layer transformer, which reconstructs the original sentence based on the [CLS] embedding. The other one is for OT decoding: it is based on a linear projection unit LPU (C), which transforms the ordinary token embeddings into the vocabulary space and aggregate them via max-pooling to preserve the BoW feature of the input. 

\section{Related Works}
%% Dense retrieval
%% PLM and retrieval
%% Representation beyond cls 

Neural retrieval is critical for many real-world applications, such as web search, question answering, advertising and recommender systems \cite{karpukhin2020dense,zhang2022uni,xiao2022training,xiao2021matching,xiao2022progressively}. It maps the query and document into embeddings within the same latent space, making their semantic relationship to be measured by the embedding similarity. In recent years, the pre-trained language models have been widely applied to deep semantic retrieval such that discriminative representations can be generated for the queries and documents. Despite the preliminary progress achieved by early pre-trained models, like BERT \cite{Devlin2019BERT}, it is noticed that the more advanced models bring little benefit to the retrieval quality, and it's believed that the conventional pre-training algorithms are not compatible with the purpose of deep semantic retrieval \cite{gao2021condenser,lu2021less,wang2022simlm}. 

To mitigate the above problem, people become increasingly interested in developing retrieval oriented pre-trained models. For example, it is proposed to leverage self-contrastive learning (SCL) where the language models are pre-trained to discriminate positive samples generated by data augmentation and in-batch negative samples \cite{chang2020pre,guu2020realm,izacard2021towards}. The SCL based algorithms are limited by many factors, like the quality of data augmentation and the requirement of huge amounts of negative samples. Later on, the auto-encoding based algorithms receive growing interests: the input sentences are encoded into embeddings, based on which the original sentences are reconstructed \cite{lu2021less,wang2021tsdae}. The recently proposed methods, such as SimLM \cite{wang2022simlm} and RetroMAE \cite{liu2022retromae}, extend the previous auto-encoding framework by upgrading the encoding and decoding mechanisms, which substantially improves the quality of deep semantic retrieval. 

The existing retrieval-oriented pre-training methods target on improving the semantic representation capacity for the contextualized embedding from the [CLS] token. However, it is noticed that the ordinary tokens may provide additional information besides [CLS], especially when dealing with long and semantic-rich documents \cite{luan2021sparse,humeau2019poly,lin2022aggretriever}. As a result, it is necessary to extend the current works, where the representation capability can be enhanced for both types of contextualized embeddings.

\section{Methodology}
We start with an overview of DupMAE in this section. The framework of DupMAE is shown as Figure \ref{fig:2}. There is an unified encoder (A), where the masked input is encoded into its contextualized embeddings. There are two decoders working collaboratively. One decoder is applied for [CLS] decoding (B): it employs a single-layer transformer, which reconstructs the original sentence based on the [CLS] embedding. The other one is used for OT decoding (C): it utilizes a linear projection unit (LPU), which transforms the ordinary token embeddings into the vocabulary space. The transformed results are aggregated by max-pooling, where the BoW feature of the input is predicted. The two decoding tasks are jointly conducted to train the encoder. The [CLS] and OT embeddings are aggregated for the final representation of the input. With proper dimension reduction, it may preserve the same computation cost of inner-product and memory footprint as one single dense vector.

% Then, we'll elabothe rate how the pre-training is conducted for the [CLS] and OT embeddings, respectively. Finally, we'll present our formulation of semantic representation. 

% DupMAE's framework is shown as Figure \ref{fig:2}. There is an unified encoder (A), where the masked input is encoded into its contextualized embeddings. There are two decoders. One is for [CLS] decoding (B): it employs a single-layer transformer, which reconstructs the original input based on the [CLS] embedding. The other one is for OT decoding (C): it utilizes a linear projection unit (LPU), which transforms the ordinary token embeddings into the vocabulary space. The transformed results are aggregated by max-pooling, where the BoW feature of the input is predicted. The two decoding tasks are jointly conducted to train the encoder. The [CLS] and OT embeddings are aggregated into one vector. With proper dimension reduction, it may preserve the same computation cost of inner-product and memory footprint. 

% The embeddings from CLS and ordinary tokens are jointly learned based on the two decoders. With reduced dimensions, the whole embeddings are aggregated to represent the input, such that the memory footprint will be similar as using one single embedding. 

% The two decoding components are jointly trained based on sampled sentences from the pre-training corpus. With reduced dimensions, the contextualized embeddings are aggregated to represent the input sentence, such that the memory footprint will be similar as one single embedding. 

% \subsection{Encoding and [CLS] Decoding}
\subsection{Encoding}
The input sentence $X$ is sampled and masked as $\tilde{X}_{enc}$ by randomly replacing some of its tokens with the special token [M]. A moderate masking ratio is applied during the encoding stage (30\%); as a result, the majority of the input information will be preserved by encoding result. The encoding network $\Phi^{enc}(\cdot)$ is used to transform the masked sentence into the contextualized embeddings for [CLS] ($\mathbf{h}_{\tilde{X}}$) and ordinary tokens ($\mathbf{H}_{\tilde{X}_{enc}}$): 
\begin{equation}
    \mathbf{h}_{\tilde{X}}, ~ \mathbf{H}_{\tilde{X}_{enc}} \leftarrow \Phi_{enc}(\tilde{X}_{enc}). 
\end{equation}
In order to capture the in-depth semantics about the sentence, a full-scale BERT-like encoding network is used to generate to the contextualized embeddings. The masked tokens for the encoder are predicted following the typical form of masked language modeling (MLM) \cite{Devlin2019BERT}. The training loss of MLM is denoted as $\mathcal{L}_{mlm}$. 

\subsection{[CLS] Decoding}
The [CLS] embedding joins with the masked input (re-generated) to decode the original sentence. Following the recent auto-encoding based pre-training methods \cite{liu2022retromae, wang2022simlm}, the decoding is performed with a simplified network and an aggressive masking ratio. 
These settings will force the embedding to fully capture the input information where high-fidelity reconstruction can be made. 
Particularly, the input $X$ is masked as $\tilde{X}_{dec}$, with half of its tokens selected for masking. 
% As for decoding, the input sentence $X$ is masked again as $\tilde{X}_{dec}$. A comparatively higher masking ratio is applied, where half of the input tokens will be masked. The masked input is joined with the sentence embedding, based on which the original sentence is reconstructed. 
An one-layer transformer is utilized for decoding, and two hidden-state streams: $\mathbf{H}_1$ (query stream), $\mathbf{H}_2$ (context stream), are used as the input: 
% RetroMAE is highlighted for its enhanced decoding, where the two-stream self-attention and the position-specific attention mask are utilized. Particularly, it generates two input streams: $\mathbf{H}_1$ (query) and $\mathbf{H}_2$ (context), for decoding:
\begin{equation}
\begin{gathered}
\label{eq:4}
\mathbf{H}_1 \leftarrow [\mathbf{h}_{\tilde{X}} + \mathbf{p}_0,...,
\mathbf{h}_{\tilde{X}} + \mathbf{p}_N], \\
\mathbf{H}_2 \leftarrow  
[\mathbf{h}_{\tilde{X}}, \mathbf{e}_{x_1}+\mathbf{p}_1, ... , \mathbf{e}_{x_N}+\mathbf{p}_N].
\end{gathered}
\end{equation}
Here, $\mathbf{h}_{\tilde{X}}$ is the [CLS] embedding from encoder, $\mathbf{e}_{x_i}$ is the $i$-th token embedding, $\mathbf{p}_i$ is the $i$-th position embedding.
Given the above input, it performs self-attention w.r.t. the mask matrix $\mathbf{M} \in \mathbb{R}^{L \times L}$:
% We introduce the position-specific attention mask $\mathbf{M} \in \mathbb{R}^{L \times L}$, where self-attention is performed as:
\begin{equation}\label{eq:5}
\begin{gathered}
    \mathbf{Q} = \mathbf{H}_1\mathbf{W}^Q, \mathbf{K} = \mathbf{H}_2\mathbf{W}^K, \mathbf{V} = \mathbf{H}_2\mathbf{W}^V; \\
    \mathbf{M}_{ij} = 
    \begin{cases}
    0, ~~~~~~\text{can be attended}, \\
    -\infty, ~\text{masked}; 
    \end{cases} \\
    \mathbf{A} = \mathrm{softmax}(\frac{\mathbf{Q}^T\mathbf{K}}{\sqrt{d}} + \mathbf{M
    })\mathbf{V}. 
\end{gathered}
\end{equation}
The output $\mathbf{A}$, together with $\mathbf{H}_{1}$ (from the residual connection) are used to predict the original input.
% (other operations, like layer-norm and FFN, are omitted from our discussion). 
Finally, the following objective is optimized: 
\begin{equation}\label{eq:6}
 \mathcal{L}_{dec} = \sum\nolimits_{x_i \in X} \mathrm{CE}(x_i|\mathbf{A}, \mathbf{H}_{1}).
\end{equation} 
As the decoder only contains one transformer layer, each token $x_i$ is reconstructed based on the unique context which are visible to the $i$-th row of $\mathbf{M}$. The mask matrix is generated by the following rules:  
\begin{equation}\label{eq:7}
\begin{gathered}
\mathbf{M}_{ij} = 
    \begin{cases}
    0,   ~~ x_j \in s(X_{\neq i}), ~\text{or}~ j_{{|i\neq0}}=0 \\
    -\infty,  ~~ \text{otherwise}.
    \end{cases} 
\end{gathered}
\end{equation}
In the $i$-th row, the sampled positions $s(X_{\neq i})$ and the first position are set to 0, meaning that they will be made visible to the $i$-th token during self-attention. Meanwhile, the non-sampled positions and the diagonal position (indicating the position of the $i$-th token itself) will be $-\infty$, which will keep them masked during self-attention.

\subsection{OT Decoding and Training Objective}\label{sec:method-dup}
% The merit of [CLS] decoding can be summarized by the following statement: because of the simplicity of decoding network, the encoder is forced to fully preserve the input information by its contextualized embeddings, such that the original input can be reconstructed with high fidelity. Given its effectiveness in pre-training [CLS] embedding, the same spirit is extended to improve the semantic representation capacity of OT embeddings. In this place, a simple and lightweight decoding workflow is designed, which consists of two basic operations shown as Figure \ref{fig:2} (C). 

% complementary
% lexical feature 

% The spirit of [CLS] decoding can be stated as follows: given the simplicity of decoder, the [CLS] embedding is forced to fully encode the input semantics, such that high-fidelity reconstruction can be produced. To extend such a decoding process for OT embeddings, two problems need to be addressed: 1) the aggregation of the whole OT embeddings, 2) the formulation of decoding objective. Here, we take inspiration from the recent works on sparse representation \cite{zhao2020sparta,formal2021splade} to aggregate OT embeddings, based on which is an even simplified decoding objective is designed. 

The decoding task for OT embeddings are designed based on two considerations. On one hand, it will follow the same spirit as the [CLS] decoding task, where the decoding network is designed to be simplified. 
% By doing so, the embeddings will be forced to fully encode the input information in order to make high-fidelity reconstruction. 
On the other hand, it will take a differentiated objective with the [CLS] decoding; therefore, it may facilitate the two types of embeddings to capture complementary information. In this place, we proposed the following decoding task for OT embeddings. 

First of all, the OT embeddings (with masked tokens excluded) $\mathbf{H}_{\tilde{X}_{enc}}$: $\{\mathbf{h}_{x_1}, ..., \mathbf{h}_{x_N}\}$ are linearly transformed into the vocabulary space: 
\begin{equation}\label{eq:8}
    \boldsymbol{\mu}_{x_i} \leftarrow \mathbf{h}_{x_i}^T\mathbf{W}^O, 
    ~ x_i \in \tilde{X}_{enc},
\end{equation}
($\mathbf{W}^O$ $\in$ $\mathbb{R}^{d \times |V|}$, $d$: embedding dimension, $|V|$: vocabulary size.) The transformed results are aggregated through token-wise max-pooling:
\begin{equation}\label{eq:9}
    \boldsymbol{\mu}_{\tilde{X}_{enc}} \leftarrow token.\mathrm{Max}(\{\boldsymbol{\mu}_{x_i}|\tilde{X}_{enc}\}),
\end{equation}
where the largest activation values of all tokens in $\tilde{X}_{enc}$ will be preserved for each vocabulary. 

% We expect the BoW feature of the input to be preserved by $\boldsymbol{\mu}_{x_i} \in \mathbb{R}^{|V|}$. As a result, the following objective $\mathcal{L}_{BoW}$ is formulated: 

Secondly, we propose the following objective where the BoW feature of the input is recovered. As a result, the lexical information can be better encoded by the OT embeddings.
\begin{equation}\label{eq:10}
    \mathrm{min}. - \sum_{x \in set(X)} \log 
    \frac{\exp(\boldsymbol{\mu}_{\tilde{X}_{enc}}[x])}
    {\sum_{x' \in V} \exp(\boldsymbol{\mu}_{\tilde{X}_{enc}}[x'])},
\end{equation}
where $x\in set(X)$ is a unique token of the input $X$, $V$ is the whole vocabulary. The encoder's loss, the decoding losses from [CLS] (Eq. \ref{eq:6}) and OT (Eq. \ref{eq:10}) are added up as our training objective: 
\begin{equation}\label{eq:11}
    \mathrm{min}. ~ \mathcal{L}_{mlm} + \mathcal{L}_{dec} + \mathcal{L}_{BoW}.
\end{equation}

\subsection{Representation}
A remaining problem of DupMAE is how to generate the semantic representation for the input. It's expected that the [CLS] and OT embeddings can be collaborated, where a stronger representation can be produced. Besides, it has to be compact, such that the retrieval process can be efficient in terms of computation cost and memory consumption. To these ends, we propose the following aggregation method. Firstly, the [CLS] embedding $\mathbf{h}_X$ is linearly transformed to a lower dimension ($d'$): 
\begin{equation}\label{eq:12} 
    \mathbf{\hat{h}}_X \leftarrow \mathbf{h}_X^T \mathbf{W}^{cls}, ~
    \mathbf{W}^{cls} \in \mathbb{R}^{d \times d'}.
\end{equation}
Secondly, knowing that the OT embeddings are aggregated into a high-dim vector $\boldsymbol{\mu}_{X}$, we directly reduce its dimension via sparsification:
% we leverage $\boldsymbol{\mu}_{X}$ to aggregate the information from OT embeddings, where we reduce its dimension via sparsification: 
% \begin{equation}\label{eq:13}
%     \boldsymbol{\hat{\mu}}_{X} 
%     \leftarrow
%     \{i: \boldsymbol{\mu}_{X}[i] ~| ~
%     \text{if}. ~ \boldsymbol{\mu}_{X}[i] \in \text{top-k}(\boldsymbol{\mu}_{X}) \}.
% \end{equation}
\begin{equation}\label{eq:13}
    \boldsymbol{\hat{\mu}}_{X} 
    \leftarrow
    \{i: \boldsymbol{\mu}_{X}[i] ~| ~ i \in 
    I_X \}.
\end{equation}
Here, $I_X$ stands for the indexes where $\boldsymbol{\mu}_{X}[i] \in \text{Top-k}(\boldsymbol{\mu}_{X})$, $k$ is the number of elements to be preserved for $\boldsymbol{\mu}_{X}$. For each document, we concatenate the dim-reduction results of [CLS] and OT embeddings as its semantic representation: $[\mathbf{\hat{h}}_X; \boldsymbol{\hat{\mu}}_{X}]$. For each query, we measure its relevance to a document based on the following form of inner-product: 
\begin{equation}\label{eq:14} 
    \langle q,d \rangle = \mathbf{\hat{h}}_q^T \mathbf{\hat{h}}_d + 
    \sum\nolimits_{I_d} \boldsymbol{\mu}_q[i] \boldsymbol{\mu}_d[i].
\end{equation} 
With proper configurations, the computation cost of inner product and memory footprint will be same as working conventional dense embeddings. 

\textbf{Fine-Tuning}. The pre-trained encoder is fine-tuned with three steps. Firstly, the contrastive learning is conducted for the in-batch negatives ($\text{IB}$): 
\begin{equation}\label{eq:15}
    \min. - \sum_{q} \log \frac{\exp(\langle q,d^+ \rangle)}
    {\sum_{d \in \{d^+, \text{IB}\} }\exp(\langle q,d \rangle)}. 
\end{equation}
Secondly, we get the ANN hard negatives  for each query based on the first-stage encoder $D^-$ \cite{xiong2020approximate}, and continue to perform contrastive learning with both hard and in-batch negatives: 
\begin{equation}\label{eq:16}
    \min. - \sum_{q} \log \frac{\exp(\langle q,d^+ \rangle)}
    {\sum\nolimits_{d \in \{d^+, D^-, \text{IB} \} }\exp(\langle q,d \rangle)}. 
\end{equation} 
Thirdly, we perform knowledge distillation: a cross-encoder is trained to discriminate the positives ($d^+$) from negatives ($d^-$) for each query. Then, the soft labeled cross-entropy is minimized: 
\begin{equation}\label{eq:17}
    \min. - \sum_q \sigma_q^d \log \frac{\exp(\langle q,d^+ \rangle)}
    {\sum_{d \in \{d^+, D^-\} }\exp(\langle q,d \rangle)}
\end{equation} 
where $\sigma_q^d$ is the softmax activation of the cross-encoder's prediction of q and d's relevance. 

The first two fine-tuning steps are cost effective, as they only involve low-cost operations. The third step will bring a much larger cost due to the training and scoring of the cross-encoder. Nevertheless, it also helps to fine-tune the model for a better precision. In our experiments, comprehensive analysis is made for DupMAE's impact on different stages.

\section{Experiment}
The empirical studies are conducted to explore the following research questions. 
\begin{itemize}
    \item \textbf{RQ 1.} Whether DupMAE produces better semantic representations, compared with the existing competitive pre-training baselines? 
    \item \textbf{RQ 2.} Whether DupMAE is able to maintain its advantages throughout different situations?
    \item \textbf{RQ 3.} Whether DupMAE benefits from the joint utilization of both [CLS] and OT embeddings, and what's the individual contribution from each embedding?
    \item \textbf{RQ 4.} Whether the pre-training tasks contribute to both [CLS] and OT embeddings?
\end{itemize}
% Empirical studies are made to explore the following research questions. \textbf{RQ 1.} Whether DupMAE gives better semantic representations, especially compared with its peer RetroMAE? \textbf{RQ 2.} Whether DupMAE maintains its advantages in different situations? \textbf{RQ 3.} Whether DupMAE benefits from the joint usage of [CLS] and OT embeddings, and what's the individual contribution from each embedding? \textbf{RQ 4.} Whether the pre-training tasks contribute to both [CLS] and OT embeddings? 

\textbf{Benchmarks}. The experiments are conducted for both supervised and zero-shot settings. We choose the \textbf{passage} and \textbf{document} retrieval task of \textbf{MS MARCO} benchmark \cite{nguyen2016ms} for supervised evaluations. It contains queries from Bing Search, where ground-truth answers to the queries need to be retrieved from 8.8 million passages and 3 million documents, respectively. The queries from the dev set and TREC Deep Learning track in 2019 (DL'19) \cite{craswell2020overview} are used for evaluation. We leverage \textbf{BEIR} benchmark \cite{thakur2021beir} for zero-shot evaluations. It contains a total of 18 datasets, which covers diverse types of retrieval tasks, such as question answering, duplication detection, and fact verification, etc. Following the official evaluation script, the pre-trained models are fine-tuned with MS MARCO queries, and evaluated for their out-of-domain retrieval performances on each of the 18 datasets.

% We choose the following datasets in our experiments, where we may evaluate DupMAE comprehensively. (1) \textbf{MS MARCO} \cite{nguyen2016ms}, which contains queries from Bing Search. We use the passage retrieval task, where passages covering the ground-truth answers need to be retrieved for each query. There are 8.8 million passages in the corpus, 502,939 queries in the training set, and 6,980 queries in the evaluation set (Dev). (2) \textbf{Natural Questions} \cite{kwiatkowski2019natural}, which contains queries from Google. There are 79,168 training queries, 8,757 dev queries, and 3,610 testing queries. The answer is retrieved from 21,015,324 Wikipedia passages \cite{karpukhin2020dense}. (3) \textbf{BEIR} \cite{thakur2021beir}, which is an authority benchmark for zero-shot retrieval. It contains 18 datasets in total, which covers different types of retrieval tasks, e.g., question answering, entity retrieval, fact verification, etc. The pre-trained models are fine-tuned with queries from MS MARCO, and evaluated for their out-domain retrieval performance on each of the 18 datasets. 

\textbf{Baselines}. We consider the following baselines for supervised evaluations according to their fine-tuning strategies. The first one only leverage \textbf{hard or in-batch negatives}, including ANCE \cite{xiong2020approximate}, SEED \cite{lu2021less}, ADORE \cite{zhan2021optimizing}, COSTA \cite{ma2022pre}, PROP \cite{ma2021prop}, B-PROP \cite{ma2021b}, Condenser \cite{gao2021condenser}, and coCondener \cite{gao2021unsupervised}. The second type leverage \textbf{sophisticated fine-tuning} strategies like {knowledge distillation}, including RocketQAv2 \cite{ren2021rocketqav2}, AR2 \cite{zhang2021adversarial}, AR2+SimANS \cite{zhou2022simans}, SPLADEv2 \cite{formal2021splade}, ColBERTv2 \cite{santhanam2021colbertv2}, ERNIE-Search \cite{lu2022ernie}, SimLM \cite{wang2022simlm}, RetroMAE \cite{liu2022retromae}. We emphasize two methods for zero-shot evaluations. One is {BM25}, which is a common sparse retrieval method and a strong baseline in zero-shot settings. The other type are the large-scale pre-trained retrievers based on contrastive learning: Contriever \cite{izacard2021towards} and the family of GTR-* \cite{ni2021large}. Among them, GTR-XXL is a super large model with 4.8B parameters (over 40$\times$ larger than BERT base). 

% The second one include dense retrievers based on \textbf{pre-trained language models}, like BERT \cite{Devlin2019BERT} and RetroMAE \cite{liu2022retromae}. Among them, Contriever \cite{izacard2021towards} and GTR-* \cite{ni2021large} are pre-trained with massive amounts of data through contrastive learning, and GTR-XXL is a super large model with 4.8 billion parameters (more than 40$\times$ larger than other methods). 

% We consider three types of baselines in our experiments. Firstly, we use generic pre-trained models, including {BERT} \cite{Devlin2019BERT}, {RoBERTa} \cite{Liu2019Roberta}, {DeBERTa} \cite{he2020deberta}, which have been widely applied for dense retrieval \cite{karpukhin2020dense}. Secondly, we use retrieval-oriented pre-trained models, including Condenser \cite{gao2021condenser}, coCondenser \cite{gao2021unsupervised}, SEED \cite{lu2021less}, SimLM \cite{wang2022simlm}, RetroMAE \cite{liu2022retromae}, whose pre-training tasks are tailored for deep semantic retrieval. Thirdly, we also consider other representative retrievers in recent years, like RocketQAv2 \cite{qu2020rocketqa}, ERNIE-Search \cite{lu2022ernie}, ColBERTv2 \cite{santhanam2021colbertv2}, Contriever \cite{izacard2021towards} and GTR \cite{ni2021large}, which achieve strong performances on our experimental datasets. 

\begin{table}[t]
    \centering
    % \small
    \scriptsize
    % \footnotesize
    \begin{tabular}{|p{2.2cm}|C{1.2cm}|C{1.2cm}|C{1.3cm}|}
    % \ChangeRT{1pt} 
    \hline
    &
    \multicolumn{2}{c|}{\textbf{Passage Dev}} & \multicolumn{1}{c|}{\textbf{DL'19}} \\
    % \cmidrule(lr){1-1}
    % \cmidrule(lr){2-3}
    % \cmidrule(lr){4-4}
    \hline
    \textbf{Methods} &
    \textbf{MRR@10} & \textbf{R@1000} & \textbf{NDCG@10} \\
    \hline
    ANCE & 0.330 & 0.959 & 0.648 \\
    SEED & 0.339 & 0.961 & -- \\
    Condenser & 0.366 & 0.974 & 0.698 \\
    coCondenser & 0.382 & 0.717 & 0.684 \\
    RocketQAv2 & 0.388 & 0.981 & -- \\
    AR2 & 0.395 & 0.986 & -- \\
    AR2+SimANS & 0.409 & 0.987 & -- \\
    SPLADEv2 & 0.368 & 0.979 & 0.729 \\
    ColBERTv2 & 0.397 & 0.984 & -- \\
    ERNIE-Search & 0.401 & 0.982 & -- \\
    SimLM & 0.411 & 0.987 & 0.714 \\
    RetroMAE (stage 3) & 0.416 & 0.988 & 0.681 \\
    % \hhline{=|=|=|=|=}
    \hline
    DupMAE (stage 2) & 0.410 & 0.987 & 0.713 \\
    DupMAE (stage 3) & \textbf{0.426}& \textbf{0.989} & \textbf{0.751} \\
    \hline
    % \ChangeRT{1pt}
    \end{tabular}
    \vspace{-5pt}
    \caption{\small{MS MARCO passage retrieval.}} 
    \vspace{-10pt}
    \label{tab:2}
\end{table}

\textbf{Implementation details}. DupMAE utilizes a bi-directional transformer network as its encoder, with 12 layers, 768 hidden-dim, and a vocabulary of 30522 tokens (same as BERT base). The decoder is an one-layer transformer. The [CLS] embedding and OT embedding are reduced to dim-384 by default. As a result, it will preserve the same computation cost of inner-product as the baselines which use dim-768 embeddings. We also explore other configurations of dimensions in our experiments. The masking ratio is set to 0.3 for encoder and 0.5 for decoder. We leverage three commonly used corpora for pre-training: Wikipedia, BookCorpus \cite{Devlin2019BERT}, and MS MARCO \cite{nguyen2016ms}. The pre-training and fine-tuning take place on machines with 8$\times$ Nvidia V100 (32GB) GPUs. The models are implemented with PyTorch 1.8 and HuggingFace transformers 4.16.

\subsection{Main Results} 
% The {supervised evaluation} results on MS MARCO and the zero-shot evaluation results on BEIR are reported in Table \ref{tab:2}, \ref{tab:2-2}, and \ref{tab:3}, respectively. 

\begin{table}[t]
    \centering
    % \small
    \scriptsize
    % \footnotesize
    \begin{tabular}{|p{2.2cm}|C{1.2cm}|C{1.2cm}|C{1.2cm}|}
    % \ChangeRT{1pt} 
    \hline
    &
    \multicolumn{2}{c|}{\textbf{Document Dev}} & \multicolumn{1}{c|}{\textbf{DL'19}} \\
    % \cmidrule(lr){1-1}
    % \cmidrule(lr){2-3}
    % \cmidrule(lr){4-4}
    \hline 
    \textbf{Methods} &
    \textbf{MRR@100} & \textbf{R@100} & \textbf{NDCG@10} \\
    \hline
    BM25 & 0.277 & 0.807 & 0.519 \\
    BERT & 0.389 & 0.877 & 0.594 \\
    ICT & 0.396 & 0.882 & 0.605 \\
    PROP & 0.394 & 0.884 & 0.596 \\
    B-PROP & 0.395 & 0.883 & 0.601 \\
    COIL & 0.397 & -- & 0.636 \\
    ANCE (first-p) & 0.377 & 0.893 & 0.615 \\
    ANCE (max-p) & 0.384 & 0.906 & 0.628 \\
    STAR & 0.390 & 0.913 & 0.605 \\
    Adore & 0.405 & 0.919 & 0.628 \\
    SEED & 0.396 & 0.902 & 0.605 \\
    % \hhline{=|=|=|=|=}
    COSTA & 0.422 & 0.919 & 0.626 \\
    RetroMAE (stage 2) & 0.432 & 0.935 & 0.593 \\
    \hline
    DupMAE (stage 2) & \textbf{0.451} & \textbf{0.950} & \textbf{0.667} \\
    % \ChangeRT{1pt}
    \hline
    \end{tabular}
    \vspace{-5pt}
    \caption{\small{MS MARCO document retrieval.}} 
    \vspace{-10pt}
    \label{tab:2-2}
\end{table}

The \textbf{supervised evaluations} are shown as Table \ref{tab:2} and \ref{tab:2-2}, where the following observations can be made. Firstly, DupMAE achieves superior performances on both tasks of MS MARCO. For passage retrieval, it reaches a MRR@10 of \underline{0.426}, outperforming the previous SOTA pre-trained models, like SimLM and RetroMAE, by \underline{+1\%} absolute point. For document retrieval, it achieves a MRR@100 of \underline{0.451}, leading to \underline{{+1.9\%}} absolute improvements. Such observations indicate that the pre-trained model's representation quality is substantially improved with DupMAE. Note that DupMAE's performances are much higher than baselines like ColBERTv2, SPLADE, and COIL. These methods utilize multi-vector for semantic representation, which is more expensive in terms of memory and computation. Besides, even with DupMAE (stage 2), which simply takes one-round of hard-negative sampling, we may outperform many of the baselines relying on sophisticated fine-tuning strategies, like knowledge distillation (ColBERTv2, ERNIE-Search) and joint learning of retriever and ranker (AR2, AR2+SimANS). 

To summary, the above observations reflect DupMAE's two-fold merits to real-world applications: \textbf{1.} it improves the best performance where neural retrievers may get, \textbf{2.} it helps to produce strong retrieval quality in a cost-effective way. 

% though knowledge distillation helps a lot with the retrieval quality, as the distillation based methods like AR2 and SimLM are much stronger than those without distillation, e.g., coCondenser, it is quite an expensive operation which call for frequent scoring of the teacher model. 

% Secondly, the retrieval quality may benefit substantially from the expensive knowledge distillation based fine-tuning, as the methods in the middle box are generally better than those in the upper box. However, we may also observe that with our enhanced pre-trained model, DupMAE (stage 2) is able to outperform the majority of the baseline methods with relatively low-cost fine-tuning operations, i.e., purely with hard negative samples. 
% The two observations jointly reflect DupMAE's value to real-world applications: it will not only improve the best performance where neural retrievers may get, but also help to produce strong retrieval quality in a cost-effective way. 

\begin{table*}[t]
    \centering
    % \small
    \scriptsize
    % \footnotesize
    \begin{tabular}{|p{1.8cm}|C{0.9cm}|C{0.9cm}|C{0.9cm}|C{0.9cm}|C{0.9cm}|C{0.9cm}|C{0.9cm}|C{0.9cm}|C{0.9cm}|C{0.9cm}|}
    % \ChangeRT{1pt}
    \hline
    \textbf{Datasets} & \Rot{\textbf{BM25}} & 
    \Rot{\textbf{BERT}} & \Rot{\textbf{SEED}} & \Rot{\textbf{Condenser}~} & \Rot{\textbf{Contriever}~} & \Rot{\textbf{GTR-base}~} & \Rot{\textbf{GTR-XXL}~} & \Rot{\textbf{RetroMAE}~} & \Rot{\textbf{DupMAE}~} & \Rot{\textbf{DupMAE}\textsuperscript{\textdagger}} \\
    % \hline
    \hhline{|=|=|=|=|=|=|=|=|=|=|=|}
    TREC-COVID & 0.656 & 0.615 & 0.627 & 0.750 & 0.596 & 0.539 & 0.501 & \underline{\textbf{0.772}} & 0.728 & {0.770}$^\uparrow$ \\
    BioASQ & 0.465 & 0.253 & 0.308 & 0.322 & 0.383 & 0.271 & 0.324 & 0.421 & \underline{0.508} & \textbf{{0.514}}$^\uparrow$ \\
    NFCorpus & 0.325 & 0.260 & 0.278 & 0.277 & 0.328 & 0.308 & 0.342 & 0.308 & \underline{0.346} & \textbf{{0.366}}$^\uparrow$ \\
    \hline
    NQ & 0.329 & 0.467 & 0.446 & 0.486 & 0.498 & 0.495 & {0.568} & 0.518 & \underline{0.570} & \textbf{{0.578}}$^\uparrow$ \\
    HotpotQA & 0.603 & 0.488 & 0.541 & 0.538 & 0.638 & 0.535 & 0.599 & 0.635 & \underline{0.681} & \textbf{{0.683}}$^\uparrow$ \\
    FiQA-2018 & 0.236 & 0.252 & 0.259 & 0.259 & 0.329 & 0.349 & \underline{\textbf{0.467}} & 0.316 & 0.345 & {0.375}$^\uparrow$ \\
    \hline
    Signal-1M(RT) & 0.330 & 0.204 & 0.256 & 0.261 & 0.199 & 0.261 & \underline{\textbf{0.273}} & {0.265} & 0.213 & 0.237$^\uparrow$ \\
    \hline
    TREC-NEWS & 0.398 & 0.362 & 0.358 & 0.376 & 0.428 & 0.337 & 0.346 & \underline{0.428} & {0.427} & \textbf{{0.433}}$^\uparrow$ \\
    Robust04 & 0.408 & 0.351 & 0.365 & 0.349 & 0.476 & 0.437 & \underline{\textbf{0.506}} & 0.447 & {0.479} & {0.503}$^\uparrow$ \\
    \hline
    ArguAna & 0.315 & 0.265 & 0.389 & 0.298 & 0.446 & {0.511} & \underline{\textbf{0.540}} & 0.433 & 0.474 & 0.465$^\downarrow$ \\
    Touche-2020 & \underline{0.367} & 0.259 & 0.225 & 0.248 & 0.204 & 0.205 & 0.256 & 0.237 & 0.343 & \textbf{{0.382}}$^\uparrow$ \\
    \hline
    CQADupStack & 0.299 & 0.282 & 0.290 & 0.347 & 0.345 & {0.357} & \underline{\textbf{0.399}} & 0.317 & 0.320 & 0.336$^\uparrow$ \\
    Quora & 0.789 & 0.787 & 0.852 & 0.853 & 0.865 & {0.881} & \underline{\textbf{0.892}} & 0.847 & 0.845 & 0.853$^\uparrow$ \\
    \hline
    DBPedia & 0.313 & 0.314 & 0.330 & 0.339 & {0.413} & 0.347 & 0.408 & 0.390 & \underline{0.418} & \textbf{{0.419}}$^\uparrow$ \\
    \hline
    SCIDOCS & 0.158 & 0.113 & 0.124 & 0.133 & \underline{\textbf{0.165}} & 0.149 & 0.161 & 0.150 & 0.153 & \textbf{{0.165}}$^\uparrow$ \\
    \hline
    FEVER & 0.753 & 0.682 & 0.641 & 0.691 & 0.758 & 0.660 & 0.740 & 0.774 & \underline{0.800} & \textbf{{0.817}}$^\uparrow$ \\
    Climate-FEVER & 0.213 & 0.187 & 0.176 & 0.211 & {0.237} & 0.241 & \underline{\textbf{0.267}} & 0.232 & 0.232 & 0.219$^\downarrow$ \\
    SciFact & 0.665 & 0.533 & 0.575 & 0.593 & 0.677 & 0.600 & 0.662 & 0.653 & \underline{0.699} & \textbf{{0.725}}$^\uparrow$ \\
    \hhline{|=|=|=|=|=|=|=|=|=|=|=|}
    AVERAGE & 0.423 & 0.371 & 0.391 & 0.407 & 0.448 & 0.416 & 0.458 & 0.452 & \underline{0.477} & \textbf{{0.491}}$^\uparrow$ \\
    % \ChangeRT{1pt}
    \hline
    \end{tabular}
    \vspace{-5pt}
    \caption{\small{Zero-shot retrieval (NDCG@10) on BEIR. {DupMAE}\textsuperscript{\textdagger} is the extended DupMAE via domain-adaptation, where $\uparrow$ indicates the improvement over DupMAE. The highest values {w.}/{w.o.} {DupMAE}\textsuperscript{\textdagger} are marked in \textbf{bold} and \underline{underlined}, respectively.}} 
    \vspace{-10pt}
    \label{tab:3}
\end{table*}

% For \textbf{zero-shot settings}, we compare DupMAE against the typical sparse retriever BM25 and other neural retrievers based on pre-trained language models. We report the retrieval performance on every single dataset, and measure the overall performance by taking the average of all 18 datasets. 

For \textbf{zero-shot settings}, we report the retrieval performance on every single dataset, and measure the overall performance by taking the average of all 18 datasets (Table \ref{tab:3}). Firstly, DupMAE achieves remarkable performance on BEIR, reaching an average NDCG@10 of \underline{0.477} in all 18 datasets. It outperforms its close peer RetroMAE on \underline{13 out of 18} datasets, and by \underline{+2.5\%} absolute point in total average. Secondly, it is known that BM25 is a strong baseline for zero-shot retrieval, which outperforms many of the existing pre-trained models on BEIR benchmark. Even for the massive-scale GTR-XXL, which uses as much as 4.8 billion parameters and huge amounts of pre-training data, it still loses to BM25 on 8 out 18 datasets. However, with DupMAE, we may outperform BM25 on \underline{15 out of 18} datasets, leading to as much as \underline{+5.4\%} absolute improvement in total average. The above performances are impressive considering that DupMAE is merely based on a BERT-base scale encoder and uses much less pre-training data compared with other strong baselines, like Contriever and GTR. 

Recently, it becomes popular to leverage domain-adaptation to improve neural retrievers' zero-shot performances \cite{xin2021zero,wang2021gpl}. In this place, we adopt a straightforward approach for domain adaptation: we continually perform DupMAE pre-training on BEIR unlabeled corpus before fine-tuning with the source domain training queries (denoted as {DupMAE}\textsuperscript{\textdagger}). Despite simplicity, this approach is surprisingly effective, as performances are improved on \underline{{16} out of 18} datasets, leading to an average NDCG@10 of \underline{0.491}. 

Given the analysis about the main experiment results in Table \ref{tab:2}, \ref{tab:2-2} and \ref{tab:3}, we may draw the following conclusions in response to \textbf{RQ 1} and \textbf{2}: 
\begin{itemize}
    \item \textbf{Con 1}. DupMAE makes large improvements over the baselines, verifying that it substantially contributes to the pre-trained model's representation capacity and transferability. 
    \item \textbf{Con 2}. DupMAE is able to maintain superior retrieval performances across different evaluation tasks on both supervised and zero-shot scenarios, which indicates DupMAE's strong usability in real-world applications.  
\end{itemize} 

\begin{table*}[t]
    \centering
    % \small
    \scriptsize
    % \footnotesize
    \begin{tabular}{|C{0.3cm}|p{2.4cm}|C{1.6cm}|C{1.6cm}|C{1.6cm}|C{1.6cm}|C{1.6cm}|}
    % \ChangeRT{1pt} 
    \hline
    & & 
    \multicolumn{5}{c|}{\textbf{MS MARCO (Passage) Dev}}  \\
    \hline
    & \textbf{Methods} & 
    \textbf{MRR@10} & \textbf{MRR@100} & \textbf{R@10} & \textbf{R@100} & \textbf{R@1000}  \\
    \hline
    \multirow{4}{*}{1.} 
    & RetroMAE & 0.3928 & 0.4032 & 0.6749& 0.9178 & 0.9849 \\
    & CLS decoding only & 0.4008 & 0.4099 & 0.6906 & 0.9229 & 0.9840   \\
    & OT decoding only & 0.4002 & 0.4092 & 0.6890 & 0.9213 & 0.9831   \\
    & CLS and OT decoding & \textbf{0.4102} & \textbf{0.4202} & \textbf{0.7049} & \textbf{0.9280} & \textbf{0.9874}  \\ 
    % \hline
    \hhline{|=|=|=|=|=|=|=|}
    \multirow{4}{*}{2.} 
    & CLS:768 & 0.3941 & 0.4040 & 0.6865 & 0.9174 & 0.9871  \\ 
    & OT:768 & 0.4019 & 0.4114 & 0.6934 & 0.9095 & 0.9814  \\ 
    % CLS:384 & -- & -- & -- & -- & --  \\ 
    % OT:384 & -- & -- & -- & -- & --   \\ 
    & CLS:384, OT:384 & \textbf{0.4102} & \textbf{0.4202} & \textbf{0.7049} & 0.9280 & 0.9874   \\ 
    & CLS:384, OT:260 & 0.4071 & 0.4171 & 0.7037 & \textbf{0.9293} & \textbf{0.9882}  \\ 
    % \ChangeRT{1pt}
    \hline
    \end{tabular}
    \vspace{-6pt}
    \caption{\small{Ablation studies: 1. impact from pre-training, 2. impact from embedding dimensions.}} 
    \vspace{-12pt}
    \label{tab:distill}
\end{table*}

% msmarco 
% baselines
% dupmae, retromae: stage 2 & 3
% partition methods by stage 2 // 3
% mrr@{10,100}, recall@{10,100,1000} 
% -----
% nq
% -----
% beir
% bert, roberta, condenser, seed, contriever, gtr-{base,xxl}, retromae
% -----
% echo rq 1 & 2
% -----
% ablation: using [cls], ot, both
% ablation: w./w.o. ot pre-training
% ablation: dimension
% echo rq 3 & 4 

\vspace{-10pt}
\subsection{Ablation Studies} 
After verifying DupMAE's overall effectiveness, it remains to figure out which factors contribute to its improvements. Thus, we perform ablation studies as Table \ref{tab:distill}. We use MS MARCO dataset for our exploration, and fine-tune the pre-trained models with hard negative samples (stage 2).  

We conduct the following two sets of experiments. Firstly, we explore \textbf{the impact from pre-training}, whose results are shown in the upper part of Table \ref{tab:distill}. Remember that DupMAE includes two decoding tasks as discussed in Section \ref{sec:method-dup}: CLS decoding and OT decoding, we make evaluations for three alternative forms accordingly. 1) CLS decoding only, where only the [CLS] embedding is pre-trained 2) OT decoding only, where only the OT embeddings are pre-trained, 3) CLS and OT decoding, which is exactly the pre-training method used by DupMAE. We also introduce RetroMAE for comparison. Although RetroMAE and ``CLS decoding only'' share the same pre-training task, their representations are generated differently, as DupMAE jointly uses [CLS] and OT embeddings.  

We may get the following observations from the experiment results. Firstly, the joint utilization of the two pre-training tasks leads to the optimal retrieval quality, where the MRR@10 grows beyond ``CLS only'' and ``OT only'' by almost +1\% absolute point. As a result, the effectiveness of jointly performing both pre-training tasks can be verified. Secondly, RetroMAE's performance is inferior to other methods, especially ``CLS pre-train only'' which share the pre-training task with it. Such an observation reveals the different capacity between the two semantic representations: DupMAE relies on the contextualized embeddings from both [CLS] and ordinary tokens, while RetroMAE only leverages the [CLS] token's embedding. 

We further explore \textbf{the impact from different semantic representations} in the lower part of Table \ref{tab:distill}). As introduced in Section \ref{sec:method-dup}, DupMAE's default semantic representation (dim-768) consists of two parts: half of its elements come from the linear projection of [CLS] embedding, while the other half come from the sparsification of OT embeddings (denoted as ``CLS:384, OT:384''). In this place, we consider two variational formulations: (1) ``CLS:768'', which directly uses the [CLS] embedding, and (2) ``OT:768'', where the top 768 elements of the OT embeddings are used for the representation of the input. According to the experiment results, the performance of ``OT:768'' is slightly better than ``CLS:768''. At the same time, ``CLS:384, OT:384'' (the default setting of DupMAE) gives rise to a better performance than both variational formulations. The above observations indicate that the contextualized embeddings from [CLS] and ordinary tokens may provide complementary information about the input data. As a result, the joint utilization of both types of embeddings is able to generate a more powerful semantic representation. 

Note that although ``CLS:384, OT:384'' preserves the same computation cost of inner-product as ``CLS:768'', it's memory cost is slightly higher than ``CLS:768'', as extra space is needed to save the indexes of OT embeddings' sparsification results. Particularly, each index will take about 15 extra bits for index storage knowing that the vocabulary space is 30522 . In this place, we introduce another variational formulation ``CLS:384, OT:260'' by further reducing the dimension of OT embeddings. As a result, it may take the same memory footprint as ``CLS:768''. It can be observed that the new combination ``CLS:384, OT:260'' still outperforms the first two variations, and maintains a similar performance as ``CLS:384, OT:384''. 

% Given the analysis about the ablation studies in Table \ref{tab:distill}, we may come to the following conclusions in response to \textbf{RQ 3} and \textbf{4}: 
Given the above analysis, we may come to the following conclusions in response to \textbf{RQ 3} and \textbf{4}: 
\begin{itemize}
    \item \textbf{Con 3}. The collaboration of [CLS] and OT embeddings brings stronger semantic representations, indicating that encoded information from the two types of embeddings are complementary to each other. 
    % The [CLS] and OT embeddings may provide complementary information to each other; the aggregation of them helps to generate stronger semantic representations. 
    \item \textbf{Con 4}. Both tasks: [CLS] and OT decoding, contribute to DupMAE; the joint conduct of both tasks leads to the optimal performance.   
\end{itemize}

%% settings: intro
%% more from OT than CLS
%% The combination gives the best
%% conclusion: complementary
%% reduce to save space cost
%% slight lower but still strong

%% Con 1. both pre-training tasks are beneficial, joint utilization 
%% Con 2. information from two embeddings are complementary, joint use

%% RQ1. DupMAE is empirically stronger, supervised setting
%% RQ2. DupMAE is empirically stronger, zero-shot setting
%% RQ3. DupMAE help to learn from few examples
%% RQ4. Contribution from each representation
%% RQ5. Contribution from pre-training tasks
%% RQ6. Exploration of accuracy and memory foot-print 

\section{Conclusion} 
This paper presents DupMAE, a new approach for retrieval-oriented pre-training, where the semantic representation capacities can be jointly enhanced for all contextualized embeddings of the language model. It employs two complementary tasks: one reconstructs the original input from the [CLS]'s embedding, the other one predicts the BoW features based on the OT embeddings. The two tasks are jointly conducted to learn an unified encoder. The two types of embeddings, with reduced dimensions, are aggregated to be a joint semantic representation. The effectiveness of our proposed method is empirically verified, where remarkable performances are achieved on MS MARCO and BEIR benchmarks throughout different situations. 

% It leverages RetroMAE's decoding task for [CLS]'s embedding and introduces a BoW-based decoding task for OT embeddings. The training losses from the two tasks are added up for a unified encoder. The two types of embeddings, after dimension reduction, are aggregated to be a joint semantic representation. The effectiveness of our proposed method is empirically verified, as remarkable retrieval performances are achieved on MS MARCO and BEIR throughout different situations. 

% \newpage

\newpage

\section*{Limitations}
Although DupMAE is to learn representation instead of generative models, it performs pre-training on open web data. Therefore, it is also subject to potential ethical and social risks, like bias, discrimination, and toxicity. Besides, DupMAE is pre-trained with comparatively limited amount of data due to the constraint on computation resources. Despite that it already achieves a promising retrieval performance at present, it remains to explore whether the performance can be further improved with the scaling up of pre-training data, by leveraging more high-quality datasets like C4 and OpenWebText.

\bibliographystyle{acl_natbib}
\bibliography{main}

\clearpage

\appendix
\section{Appendix}
% To demonstrate the differentiation between [CLS] and OT embeddings
% a glimpse the differentiation, understand why its useful to combine, and why the pre-training tasks make sense.
% Let each embedding work alone, success if the ground-truth answers are retrieved within the top-10 candidates. 
% Two tables: sucessful cases from [CLS] and OT
% [CLS]: accurately capture fine-grained semantic relationships. 
% OT: accurately capture the lexical similarity. 

According to our experimental results in Table \ref{tab:distill}, the [CLS] and OT embeddings may jointly produce a stronger semantic representation to improve the retrieval quality. In this place, we provide a case analysis as Table \ref{tab:appendix 1} and \ref{tab:appendix 2}, which will visualize the benefit introduced by each type of embedding, and help to explain the design of the pre-training tasks. 

\subsection{Settings}
In our exploration, the [CLS] embedding and OT embeddings (aggregated and sparsified in the same way as introduced in Section \ref{sec:method-dup}) are used independently for the retrieval tasks. That's to say, the query and answer's relationships are measured by the [CLS] embeddings' similarity and OT embeddings' similarity, respectively. We select queries from the evaluation set of MS MARCO for demonstration. For each query, we count it as a successful case w.r.t. a specific type of embeddings, if its ground-truth answer can be retrieved within the Top-10 results. If the ground-truth answer is missed by one type of embeddings, its Top-1 retrieved answer will be posted for comparison. 

\subsection{Analysis}
Given the limitation of space, we select four representative queries for demonstration. The four queries can be partitioned into two sets: in Table \ref{tab:appendix 1}, the ground-truth answers are retrieved by [CLS] embeddings; while in Table \ref{tab:appendix 2}, the ground-truth answers are retrieved by OT embeddings.  

$\bullet$ \textbf{Good cases by [CLS] embeddings}. In Table \ref{tab:appendix 1}, the two queries' ground-truth answers are retrieved by the{ [CLS] embeddings}. For both cases, it calls for the pre-trained model to capture fine-grained \textbf{semantic relationships} between the query and answer. In particular, the first query is essentially about the car brands which belong to Ford. The [CLS] embedding successfully establish the connection between ``build'' and ``own'' (marked in blue). Therefore, the ground-truth answer can be successfully retrieved. Similarly, the second query emphasizes ``cncellation'' fee. By identifying the relationship between ``cncellation'' and ``Cancel'' (marked in blue), the ground-truth answer is successfully retrieved once again. Comparatively, although OT embeddings retrieve answers with close lexical features, e.g., ``built'', ``fee'' (marked in red), they appear to be less proficient in capturing the semantic relationships in both cases, where the correct answers are missed from their top-10 results. 

$\bullet$ \textbf{Good cases by OT embeddings}. In Table \ref{tab:appendix 2}, the two queries' ground-truth answers are retrieved by the {OT embeddings}. For both cases, it calls for the pre-trained model to precisely identify the ground-truth answers, which are not only semantically close to the queries, but also contain specific \textbf{lexical features}. Particularly, the first query asks about a certain type of material called ``copper coated carbon rods''. As a result, it is important to retrieve the answer which contain exactly the same term. The [CLS] embedding finds ``copper-clad steel'' (marked in red). Although similar, it is different from the required term. While with the OT embeddings, the ground-truth answer is successfully retrieved. Note that it's challenging for this case, knowing that the related term ``Copper coated carbon electrods'' (marked in blue) is wrapped in a long passage. The second query asks about the colour which represents selflessness. Although the [CLS] embedding finds the passage which is relevant to the symbolic meaning of colour (marked in red), it ignores the key term ``selflessness'' (marked in blue). On top of the OT embeddings, it successfully retrieves the ground-truth answer, which is not only semantically close to the required topic (color symbolism), but also contains the required term (selflessness). 

% differentiated tasks
% more semantic
% more lexical
% optimize the quality of joint representation 

$\bullet$ \textbf{Discussions}. It is known that both semantic and lexical features are important to information retrieval problems, such as search engine and question answers. From the above analysis, we may observe that the two types of embeddings may have their own advantages: the [CLS] embeddings tend to be more proficient in capturing the semantic closeness, while the OT embeddings may better leverage the lexical similarity. In DupMAE, we design two differentiated auto-encoding tasks for [CLS] and OT embeddings. Although both tasks help to better encode the semantic information with the contextualized embeddings, the OT decoding task emphasizes more of the lexical information, because the BoW feature needs to be directly predicted by the aggregation results of OT embeddings. By having such differentiated tasks, the two types of embeddings may focus on strengthening their unique advantages. Finally, it will help to optimize the quality of the joint representation when both types of embeddings work collaboratively. 

\begin{table*}[t]
    \centering
    % \small
    \scriptsize
    % \footnotesize
    \begin{tabular}{|p{3cm}|p{6cm}|p{6cm}|}
    % \ChangeRT{1pt} 
    \hline
    \textbf{Query} & \textbf{Retrieved answer by [CLS] embedding} & \textbf{Retrieved answer by OT embeddings}  \\
    \hline
    what cars does ford \blue{build}?
    & What car companies does Ford \blue{own}? Ford owns Jaguar (-30\%), Land Rover (-50\%), Aston Martin (-\%10), Lincoln, Mercury, Volvo (-70\%), and Mazda (-40\%). I'm not quite sure of those percentages, nor am I sure if Ford owns 100\% owns Lincoln and Volvo, but there's the basic gist of what Ford owns now. The above answer is incorrect. Ford has sold Jaguar, Volvo, \& Land Rover. (\textbf{Ground-Truth. Rank 4th})
    & Passenger Cars. The Taurus, Sable and Lincoln are \red{built} in Chicago, while many of Ford's engines are assembled in Brook Park, Ohio, with one Dearborn, Michigan, plant dedicated solely to auto parts. (\textbf{Rank 1st})
    \\
    \hline
    delta airlines \blue{cncellation} fee?
    & How to \blue{Cancel} Flights on Delta Air Lines. When the credit is used to pay for new flights, the change fee will be assessed. For example, say you bought non-refundable Delta domestic flight tickets for \$650, but your plans changed. When you are ready to purchase new flights, the fare has increased to \$700. 
    Your credit is \$650 âx80x93 \$200 change fee = \$450, so your out-of-pocket cost to buy the new ticket is \$700 - \$450 = \$250. 
    Make sure to inform Delta before departure that you will not be on the flight and request the travel credit.our credit is \$650 âx80x93 \$200 change fee = \$450, so your out-of-pocket cost to buy the new ticket is \$700 - \$450 = \$250. 
    Make sure to inform Delta before departure that you will not be on the flight and request the travel credit. 
    (\textbf{Ground-Truth. Rank 3rd})
    & As of publication, Delta charges a minimum \red{fee} of \$178 for most domestic flights and \$250 on flights to Alaska, Hawaii and the Virgin Islands, with additional charges based on the pet and carrier weight. (\textbf{Rank 1st})
    \\
    \hline
    \end{tabular}
    % \vspace{-5pt}
    \caption{Cases where the [CLS] embedding helps to retrieve the ground-truth answers.} 
    % \vspace{-5pt}
    \label{tab:appendix 1}
\end{table*} 

\begin{table*}[t]
    \centering
    % \small
    \scriptsize
    % \footnotesize
    \begin{tabular}{|p{3cm}|p{6cm}|p{6cm}|}
    % \ChangeRT{1pt} 
    \hline
    \textbf{Query} & \textbf{Retrieved answer by [CLS] embedding} & \textbf{Retrieved answer by OT embeddings}  \\
    \hline
    what are \blue{copper coated carbon rods} used for?
    & \red{Copper-clad} steel (CCS), also known as \red{copper-covered} steel or the trademarked name \red{Copperweld} is a bi-metallic product, mainly used in the wire industry that combines the high mechanical resistance of steel with the conductivity and resistance to corrosion of copper. (\textbf{Rank 1st})
    & Coidan Graphite Products supply Graphite Electrodes primarily used for the secondary production of steel EAF and ladle furnaces. Our graphite electrode stock has additional applications, such as melting products in smelting furnaces, non-ferrous metals, ceramic products and to recycle waste. There are several grades of graphite electrodes, we can match the grade with the application to lower your melting costs. Please click through to see properties of the graphite electrodes we can offer, RP grade, HP grade, SHP grade and UHP graphite electrodes. In addition we supply graphite EDM electrodes for the mould makers together with many other Spark Erosion applications. \blue{Copper coated carbon electrodes} of many shapes and sizes are used as gouging rods and welding rods in foundry applications. (\textbf{Ground-Truth. Rank 8th})
    \\
    \hline
    what color represents \blue{selflessness}?
    & But since it is also taken as off-white, it can be the color of degradation or cowardice. Orange. \red{Symbolic} of endurance and strength, orange is the color of fire and flame. it represents the red of passion tempered by the yellow of wisdom. It is the \red{symbol} of the sun. (\textbf{Rank 1st})
    & Color Symbolism - The Deeper Meaning of Blue, Blue is on the visual level a calm and peaceful color. We think of it in terms of water, sky and universe. For most of us, sky and water give us a sense of familiarity and consequently of security. For many, the universe represents a larger unity and religion. Therefore, this hue expresses security and spiritual devotion. It is the color that leads to introspection and to our very essence. It represents such ideals as \blue{selflessness}, sympathy, kindness, compassion and dedication. Blue is assigned to the physical body and, on a larger scale, represents the material aspects of life including the planet earth. (\textbf{Ground-Truth. Rank 1st})
    \\
    \hline
    \end{tabular}
    % \vspace{-5pt}
    \caption{Cases where the OT embeddings help to retrieve the ground-truth answers.} 
    % \vspace{-15pt}
    \label{tab:appendix 2}
\end{table*}

\end{document}